\documentclass{article}
\usepackage{ijcai22}

\usepackage{times}
\usepackage{soul}
\usepackage{url}
\usepackage[hidelinks]{hyperref}
\usepackage[utf8]{inputenc}
\usepackage[small]{caption}
\usepackage{graphicx}
\usepackage{subfigure}
\usepackage{amsmath}
\usepackage{amsthm}
\usepackage{booktabs}
\usepackage{algorithm}
\usepackage{algorithmic}
\usepackage{amssymb}
\newcommand{\citet}[1]{\citeauthor{#1} [\citeyear{#1}]}

\title{Meta Reinforcement Learning with Successor Feature Based Context}

\author{
Xu Han \And Feng Wu \\
\affiliations{School of Computer Science and Technology, \\ 
University of Science and Technology of China}
\emails{hxu127@mail.ustc.edu.cn, wufeng02@ustc.edu.cn}
}

\begin{document}

\maketitle

\begin{abstract}
Most reinforcement learning (RL) methods only focus on learning a single task from scratch and are not able to use prior knowledge to learn other tasks more effectively. Context-based meta RL techniques are recently proposed as a possible solution to tackle this. However, they are usually less efficient than conventional RL and may require many trial-and-errors during training. To address this, we propose a novel meta-RL approach that achieves competitive performance comparing to existing meta-RL algorithms, while requires significantly fewer environmental interactions. By combining context variables with the idea of decomposing reward in successor feature framework, our method does not only learn high-quality policies for multiple tasks simultaneously but also can quickly adapt to new tasks with a small amount of training. Compared with state-of-the-art meta-RL baselines, we empirically show the effectiveness and data efficiency of our method on several continuous control tasks.
\end{abstract}

\section{Introduction}

Reinforcement Learning (RL) has been successfully applied to a series of complex decision-making and control tasks \cite{mnih2015human,silver2017mastering}. However, the agent usually needs to go through many trial-and-error processes with the environment to learn how to take the best action. When new tasks are encountered, traditional RL may need to start learning another policy from scratch \cite{sutton2018reinforcement}. In contrast, humans and animals can learn skills quickly from a small number of examples. Indeed, they know how to use prior knowledge when adapting to new tasks. If an agent had similar capabilities, it can use information learned in previous tasks to effectively master various skills and quickly adapt to new environments. In real-world applications, many tasks often have similar internal structures \cite{finn2017model,rakelly2019efficient}. By borrowing ideas from general meta learning \cite{schmidhuber1987evolutionary,bengio1992optimization}, meta-RL methods have been proposed recently \cite{finn2017model,gupta2018meta,wang2016learning}, which aim to acquiring such common knowledge from the previous experiences of other tasks and adapting to new tasks with only a small amount of rollouts.

To date, recurrent or recursive meta-RL methods \cite{duan2016rl,wang2016learning,mishra2017simple} and gradient-based meta-RL \cite{finn2017model,rothfuss2018promp,liu2019taming} try to learn general model initialization using on-policy meta-training, which is usually sample inefficient. Context-based methods \cite{rakelly2019efficient,fakoor2019meta} are proposed to alleviate this, by learning policies with contextual information and off-policy data. By doing so, the agent can infer latent contextual variables through a small number of demonstrations or interactions with the environment in order to adapt to new tasks. In more details, it allows the agent to learn multiple tasks at the same time and easily restore the learned skills for new tasks.

For context-based meta-RL, it is crucial to learn effective latent context variables \cite{fu2020towards}. Previous work on learning context variables \cite{fakoor2019meta,rakelly2019efficient} is rather straightforward, where context variables are obtained by feeding transitions into deterministic encoders or inference networks. However, there is overlap between transitions for different tasks, which requires them to collect more training data for correcting the issue overtime \cite{li2019multi}. Recently, progress has been made by improving the context calculation methods \cite{lee2020context,li2019multi}. Unlike the existing techniques that require a lot of extra calculations, we leverage the successor feature (SF) and reward weights  from the successor features framework \cite{dayan1993improving,barreto2017successor} to improve the accuracy of the context. Note that the successor feature framework is usually used to naturally decouple the dynamics of the environment from the rewards, which makes them particularly suitable for task inference.

In this paper, we propose a new approach named meta-RL with Successor Feature based Context (SFC).  Specifically, we first uniformly sample tasks from the task distribution and use some RL method to train these tasks, in order to quickly obtain task-related knowledge. Then, we train the successor feature network. In more details, we input the transition into this network and get the corresponding successor features and reward weights. Given this, the reward structure and environment dynamics are decomposed. As successor features and reward weights contain information about environment dynamics and rewards, they are used to calculate the context variables, on which the policy is conditioned. When calculating approximating reward weights and context, we also use metric learning to improve task inference capabilities. By comparing with leading meta-RL baselines, we empirically show the effectiveness and data efficiency of our method on several continuous control tasks.

\section{Background}

We consider the following RL problem: Given that an agent has learned some skills, how can it quickly adapt to other related tasks using experiences from the learned tasks?

As aforementioned, context-based meta-RL \cite{fakoor2019meta,mishra2017simple} aims to learning multiple tasks in the meta-training stage with a context variable. Here, the context variable maps the information of past trajectories to a specific task. As a result, the meta-RL policy that chooses actions conditioned on the context variable will maximize the expected return of the corresponding task.

Intuitively, the quality of this context variable has a significant impact on sample efficiency. The context encoder to generate this variable will be challenging to train when similar transitions are gathered from different tasks \cite{li2019multi}.

\subsection{Meta-RL}

In meta-RL \cite{finn2017model}, we assume a distribution of tasks $p(\mathcal{T})$, where each task $\mathcal{T}_i$ can be modeled as a Markov decision process (MDP): $\langle \mathcal{S},\mathcal{A}, p_i, r_i, \gamma, H \rangle$. Here, $\mathcal{S}$ and $\mathcal{A}$ are the state and action spaces respectively. For each $s \in \mathcal{S} $ and $a \in \mathcal{A}$, $p_i(\cdot |s, a)$ specifies the next-state distribution after taking action $a$ in state $s$ (a.k.a. $\textit{environment dynamics}$). The reward received at transition $s \stackrel{a}{\rightarrow} s'$ is given by $r_i(s, a)$. Then, $\gamma$ is the discount factor and $H$ is the horizon.

Here, we are interested in the common meta-RL setting where all tasks share the same MDP model, except for their reward function. Given this, the set of tasks can be rewritten as: $\mathcal{T}_i = \{ \langle \mathcal{S},\mathcal{A}, p_i, r_i, \gamma, H \rangle | r_i : S \times A {\rightarrow}  \Re \}$.

Now, our objective is as follow: Given a set of training tasks $\{\mathcal{T}_1, \cdots , \mathcal{T}_K \}$ sampled from $p(\mathcal{T})$, we want to learn a multi-task policy $\pi_{\theta}(s, z)$ conditioned on the state $s$ and the context variable $z$ in the training phase. At the execution time, the agent is able to adapt to a new task $\mathcal{T}_j$ sampled from $p(\mathcal{T})$ by computing the specific context variable $z_j$ for this task using the policy $\pi_{\theta}(s, z_j)$.

\subsection{Successor Features}

Successor Representation (SR) \cite{dayan1993improving} was originally introduced for separating environment dynamics from rewards in MDPs. It allows us to dynamically compute the task-agnostic successor features and dynamic-agnostic reward weights, which is useful for task inference.

Specifically, the SR is the expected discounted future state occupancy defined as below:
\begin{equation}
	\psi^{\pi}(s, s', a)= \mathbb{E}_{\pi}\bigg[\sum_{t=0}^\infty \gamma^t\mathit{1}[s_t=s']|s_0=s, a_0 = a \bigg]
\end{equation}
where $\mathit{1}[\cdot]$ equals 1 if the argument is true and 0 otherwise.

Given policy $\pi$, starting state $\textit{s}$ and action $\textit{a}$, SR denotes the expectation of times that the state $s'$ will be visited in the future. Note that this only depends on the transition dynamics of the MDP and the policy $\pi$. With the SR, the Q function of the MDP can be represented as:
\begin{equation}
	\mathit{Q}^\pi(s,a)= \sum_{s'}\psi^{\pi}(s, s', a)\mathit{R}(s,a)
\end{equation}
As a result, the environment dynamics, captured by $\psi$, and the reward structure, modeled by $\mathit{R}$, are separated.

Recently, the SR has been naturally extended to the deep setting known as Successor Feature (SF)  \cite{barreto2017successor}, which can be applied to continuous state and action spaces straightforwardly. The basic assumption of SF is that the reward function can be parameterized as:
\begin{equation}
	R(s, a) = \phi(s, a)^\top\omega,
\end{equation}
where $\phi(s, a)$ is feature vector and $\omega$ are reward weights.

For simplicity, we rewrite the feature in an expectation form: $\phi(s) \equiv \mathbb{E}[\phi(s_0, a_0) | s_0 = s ] $. Given state $s$, the SF specifies the expected discounted features of the future as:
\begin{equation}
\psi^{\pi}(s) = \mathbb{E}_{\pi}\bigg[\sum_{t=0}^\infty \gamma^{t-1} \phi_{i+1} | s_0 = s \bigg],
\end{equation}
As mentioned above, the SF encodes information about policy $\pi$ and environmental dynamics $p(\cdot |s, a)$, and is independent of the reward function $r$ in the MDP.

\begin{figure}[t]
\centering
\includegraphics[scale=0.22]{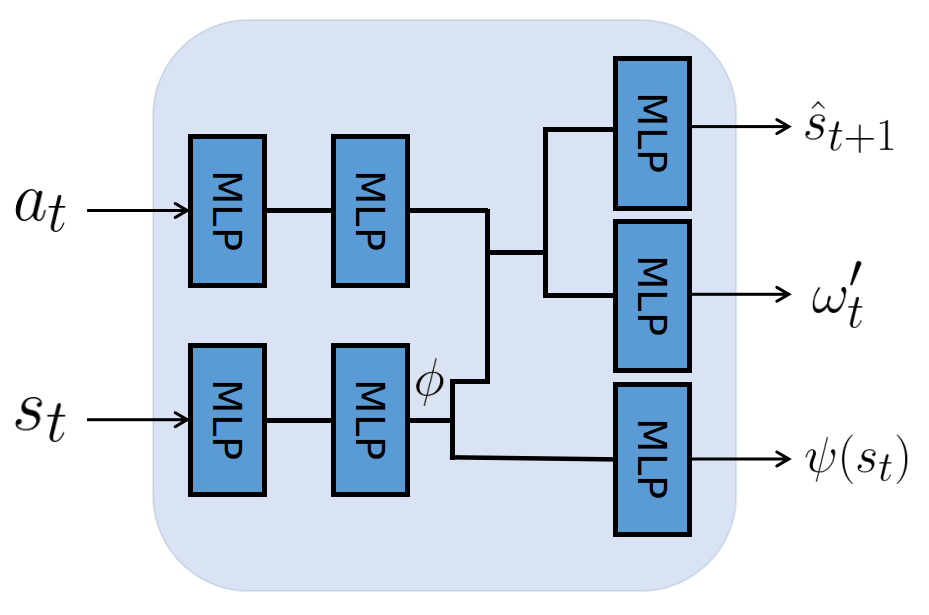}
\caption{Structure of successor feature network.}
\label{fig:1}
\end{figure}

\begin{algorithm}[t]
\caption{Training procedures of our meta-RL method}
\label{alg:algorithm1}
\begin{algorithmic}
\STATE \textbf{Input}: A set of tasks $\{\mathcal{T}_1, \cdots, \mathcal{T}_K\}$ sampled from $p(\mathcal{T})$
\STATE \textbf{Output}: The SF network $\zeta$, policy $\pi_{\theta}$, and encoder $g_{\eta}$
\STATE \texttt{// Stage 1: Collecting transitions.}
\STATE Train policies $\{ \pi_1(s), \cdots, \pi_K(s)\}$ with single-task RL
\STATE Collect corresponding transitions $\{\mathcal{D}_1, \cdots, \mathcal{D}_K\}$
\STATE \texttt{// Stage 2: Training SF network}
\STATE Randomly initialize SF network $\zeta$
\WHILE{$not\ done$}
\STATE Sample transitions from $\mathcal{D}_i$, $\forall i \in 1..K$
\STATE Calculate $\mathcal{L}_{MMD}$ by Equation \ref{Loss_MMD}
\STATE Calculate $\mathcal{L}_r , \mathcal{L}_{Recons},\mathcal{L}_{TD}$ by Equations \ref{Loss_r}, \ref{loss_recon}, and \ref{Loss_td}
\STATE Calculate $\mathcal{L}_{SF}$ by Equation \ref{Loss_SF}
\STATE Update SF network $\zeta$ by minimizing $\mathcal{L}_{SF}$
\ENDWHILE
\STATE \texttt{// Stage 3: Learning policy.}
\STATE Randomly initialize policy $\pi_{\theta}$ and context encoder $g_{\eta}$
\WHILE{$not \ done$}
\STATE Sample state-action pairs $\{(s_t, a_t)\}_i$ and transitions $\{(s_j, a_j,r_j,s_j')\}$ from $\mathcal{D}_i$, $\forall i\in 1..K$
\STATE Calculate $\{(\psi_j, \omega_j)\}_i$ using SF network $\zeta$
\STATE Calculate $\{z\}_i$ using encoder $g_{\eta}$ with $\{(\psi_j, \omega_j)\}_i$
\STATE Calculate $\mathcal{L}_{{\pi}_{\theta}}, \mathcal{L}_{MMD2}$ by Equations \ref{Loss_mse}, \ref{Loss_mmd2}
\STATE Calculate $\mathcal{L}_{IL}$ by Equation \ref{Loss_il}
\STATE Update policy $\pi_{\theta}$ and encoder $g_{\eta}$ by minimizing $\mathcal{L}_{IL}$
\ENDWHILE
\end{algorithmic}
\end{algorithm}

\begin{algorithm}[t]
\caption{Testing procedures of our meta-RL method}
\label{alg:algorithm2}
\begin{algorithmic}
\STATE \textbf{Input}: New task $\mathcal{T}_j$ sampled from $p(\mathcal{T})$ for testing
\STATE Initialize context transitions $\mathcal{D}^{\mathcal{T}_j} = \emptyset$
\FOR{$k = 1 \cdots K$}
\STATE Sample transitions $\{(s_i, a_i,r_i,s_i')\}_k$ from $\mathcal{D}_k$
\STATE Calculate $z$ using the sampled transitions
\STATE Roll out policy $\pi_{\theta}(a|s,z)$ in $\mathcal{T}_j$ to collect transitions $\mathcal{D}_k^{\mathcal{T}_j} = \{(s_l, a_l, r_l, s_l')\}$ and store $\mathcal{D}_k^{\mathcal{T}_j}$ into $\mathcal{D}^{\mathcal{T}_j}$
\ENDFOR
\STATE Choose the trajectory with maximum return in $\mathcal{D}^{\mathcal{T}_j}$
\STATE Calculate a new $z'$ using the trajectory
\STATE Evaluate policy $\pi_{\theta}(a|s,z')$
\end{algorithmic}
\end{algorithm}

\section{Algorithm}

We propose the meta-RL method with successor features context. As outlined in Algorithm \ref{alg:algorithm1}, our training procedure consists of three main stages.

In the first stage, we uniformly sample a set of $K$ tasks $\{\mathcal{T}_1, \cdots , \mathcal{T}_K \}$ from $p(\mathcal{T})$. Next, we use a vanilla single-task RL method (e.g., TD3 \cite{fujimoto2018addressing}) to learn the corresponding policies $\{ \pi_1(s), \cdots, \pi_K(s)\}$, one for each task. Then we use these policies to collect a set of transition samples for every task as $\mathcal{D} = \{\mathcal{D}_1, \cdots, \mathcal{D}_K\}$, which will be used as inputs for the following stages. In the second stage, we train a SF network that can convert the collected transitions $\mathcal{D}$ into the corresponding successor features and reward weights: $\{ (\psi_i, \omega_i') \}_{i = 1 \cdots K}$. In the third stage, we compute a context variable $z$ that depends on $\{ (\psi_i, \omega_i') \}_{i = 1 \cdots K}$ and train a policy conditioned on the context. 

Note that the second and third stages are the key steps of our approach, detailed in the following sections.

\subsection{Training Successor Feature Network}

In the original SF framework, each task corresponds to a vector of reward weight $\omega$. This vector-based representation is inefficient for tasks with high-dimensional continuous state-action spaces. Therefore, we train an encoder to output approximate reward weights $\omega'$, which has the same property as the reward weights $\omega$: $R(s, a) = \phi(s, a)^\top\omega'$. In other words, different tasks will correspond to different distributions of the approximate reward weights.

The SF network architecture that we used in this paper is shown in Figure \ref{fig:1}. The main challenge of training the SF network is to obtain an appropriate feature representation $\phi(s)$, which is the basis of the SF network outputs. Firstly, it should be a good predictor for the immediate reward given the current state as $\phi(s)$ is used to predict the reward weights $\omega'$. Secondly, it is also used to predict the next state $\hat{s}_{t+1}$ with action $a$ due to the assumption that $\phi(s)$  should also provide features that capture latent factors of the states. Note that we use the SF to ensure that there will not be much difference between similar states. Once $\phi(s)$ is learned in the network, we then compute the reward weights $\omega'$ and SF $\psi(s_t)$.

With the aforementioned objectives, the following loss functions are designed to train the SF network. Specifically, we use the Mean Squared Error (MSE) to measure the average squared difference between the predicted values and the actual values. Formally, the MSE loss is computed as:
\begin{equation}
	MSE(\ \hat{Y}, \ Y\ ) = \frac{1}{n} \sum_{i=1}^n (\hat{Y_i} - Y_i )^2
\end{equation}
where $Y_i$ is the vector of true values and $\hat{Y_i}$ is the predicted values. To learn the reward weights, the loss that we minimize for this part of the network is as:
\begin{equation}
\label{Loss_r}
	\mathcal{L}_r = MSE(\ \phi^\top\omega', \quad r\ )
\end{equation}
where $\phi^\top\omega'$ is the predicted reward and $r$ is the actual reward.

As $\phi(s)$ is the state feature representation with all the information on state $s$, we set the auxiliary goal of predicting the next state $\hat{s}_{t+1}$. Intuitively, this will make the state feature representation more robust, which is depicted as the top layers in Figure \ref{fig:1}. The loss that we minimize for this part of the network is as:
\begin{equation}
\label{loss_recon}
	\mathcal{L}_{Recons} = MSE(\ \hat{s}_{t+1}, \quad s_{t+1} \ )
\end{equation}
where $\hat{s}_{t+1}$ is the predicted state and $s_{t+1}$ is the actual state.

Similar to Q value function, SF satisfies a Bellman equation where $\phi$ play the role of rewards \cite{dayan1993improving,barreto2017successor}. Thus, the SF can actually be learned by the Temporal-Difference (TD) method, and trained by minimizing the TD loss as:
\begin{equation}
\label{Loss_td}
	\mathcal{L}_{TD} = \mathbb{E}\big[MSE(\ \phi(s_t) + \gamma \psi(s_{t+1}), \quad \psi (s_t)\ ) \big]
\end{equation}

Note that we assume that different tasks correspond to different distributions of the approximate reward weights. Given this, we leverage the Maximum Mean Discrepancy (MMD) loss borrowed from the transfer learning community to learn the approximate reward weights more robustly. Specifically, given samples $x 1,\cdots,x n \sim X$ and $y 1,\cdots,y n \sim Y$, the sampled MMD between $X$ and $Y$ is give:
\begin{equation}
\begin{split}
  MMD^2(\ X, \ Y\ ) =  \frac{1}{n^2}\sum_{i,i'}k(x_i, x_{i'}) \\
  - \frac{2}{nm}\sum_{i,j}k(x_i, y_j) \\
  + \frac{1}{m^2}\sum_{j,j'}k(y_j, y_{j'})
\end{split}
\end{equation}
where $k(\cdot, \cdot)$ is the Gaussian kernel. Let $K$ be the number of training tasks and the overall MMD loss is written as:
\begin{equation}
\label{Loss_MMD}
	\mathcal{L}_{MMD} = - \sum_{i=1}^K\sum_{j\neq i}^K MMD^2(\ \Omega_i, \ \Omega_j\ ).
\end{equation}
where $\Omega_i$ is the set of approximate reward weights in task $i$.

Now, the overall loss minimized for the SF network is as:
\begin{equation}
\label{Loss_SF}
	\mathcal{L}_{SF} = \mathcal{L}_r + \mathcal{L}_{Recons}+ \mathcal{L}_{TD} + \mathcal{L}_{MMD}
\end{equation}
where all the loss functions mentioned above are combined.

\begin{figure}[t]
\centering
\includegraphics[scale=0.25]{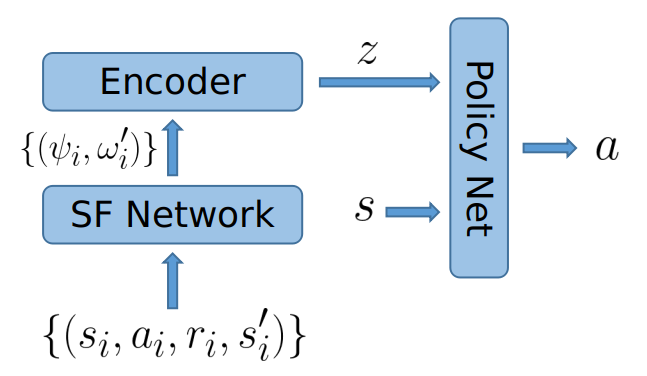}
\caption{Structure of action selection network.}
\label{fig:2}
\end{figure}

\subsection{Learning Policy with Context Variable}

We leverage a novel latent context variable $z$ depending on SFs and reward weights to learn the multi-task policy. It contains more information about environment dynamics and reward structures than the recurrent context variable \cite{fakoor2019meta,rakelly2019efficient}, directly calculated using only the state-action transitions $\{(s_i,a_i,r_i,s'_i)\}$.

In the training stage, we first sample transitions $\{ D_i \}$ from each task $i$ and feed them into the SF network to compute SFs and reward weights $\{(\psi(s)_i, \omega_i') \}$. Then, the  SFs and reward weights are put into a context encoder $g_{\eta}$, implemented by Gated Recurrent Unit (GRU) \cite{cho2014learning},  where the output $z$ of $g_{\eta}$ is set to the hidden state of the GRU.

As aforementioned, we want to learn a multi-task policy conditioned on the context variable $z_t$ for every state $s_t$, i.e. $\pi(s_t,z)$. We denote the transitions in $D_i$ as $\{ (s_1, a_1,s'_1,r_1),\cdots (s_T, a_T,s'_T,r_T) \}_i$ and use the MSE loss to minimize the action difference as:
\begin{equation}
\label{Loss_mse}
	\mathcal{L}_{MSE}=\sum_{{\tau}\sim D} \sum_{t} MSE(\ {\pi}_{\theta}(s_t, z), \ a_t\ )
\end{equation}

Similar to the first stage, we use the MMD loss for content variable $z$ in different tasks as:
\begin{equation}
\label{Loss_mmd2}
	\mathcal{L}_{MMD2} = \sum_{i=1}^K\sum_{j\neq i}^K MMD^2(Z_i, Z_j).
\end{equation}
where $Z_i$ is the set of latent context variables $z$ in task $i$.

Combining both of the losses above, the overall loss minimized for training the policy network is given as:
\begin{equation}
\label{Loss_il}
	\mathcal{L}_{IL} = \mathcal{L}_{MSE} + \mathcal{L}_{MMD2}
\end{equation}

\subsection{Adapting Policy to New Tasks}

To test our policy on a new task, we first obtain the $K$ context variables $\{z_i\}_{i=1,\cdots, K}$ based on the transition data $\{\mathcal{D}_i\}_{i=1,\cdots,K}$ stored in the training phase. This is done by sampling transitions from $\mathcal{D}_i$, feeding them to the SF network, and finally getting $z_i$ from the context encoder. Notice that this process is independent of the new task and therefore the context variables can be computed offline.

Given $\{z_i\}_{i=1,\cdots, K}$, we rollout policy $\pi_{\theta}(a|s, z_i)$ in the new task to collect a set of trajectories. we select the trajectory with maximum return to obtain a new context variable $z'$ specific to the new task. Finally, we get the evaluation result with policy $\pi_{\theta}(a|s, z')$ on the new task.

We use this method to efficiently explore the new task and compute the context variable. This can be used to adapt the learned policy to the new task. The main testing procedures of our approach are shown in Algorithm \ref{alg:algorithm2}.

\section{Experiments}

\begin{figure*}[t]
\centering
\subfigure[Ant-Fwd-Back]{
\begin{minipage}[t]{0.25\linewidth}
\centering
\includegraphics[scale=0.32]{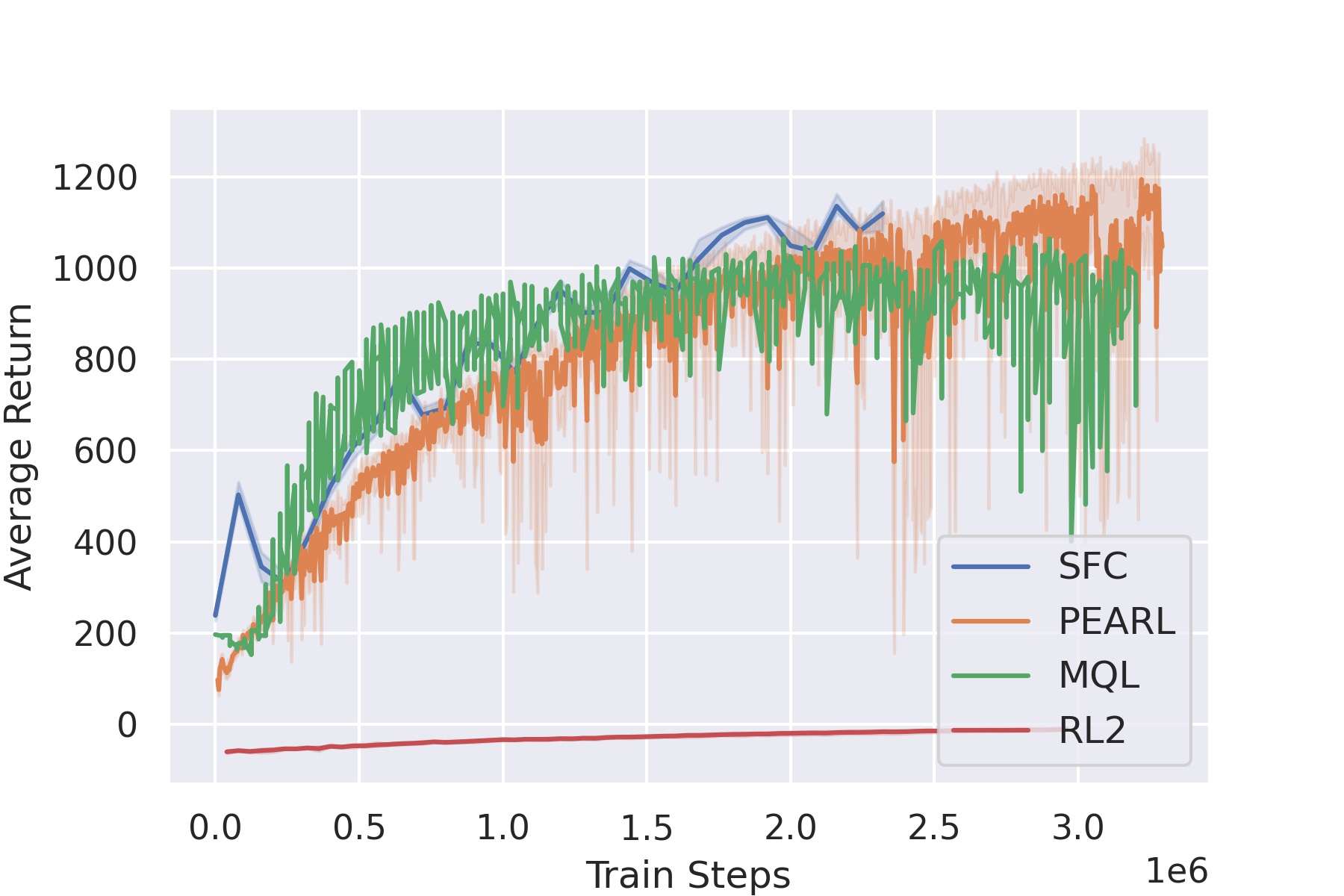}
\end{minipage}%
}%
\subfigure[Ant-Goal]{
\begin{minipage}[t]{0.25\linewidth}
\centering
\includegraphics[scale=0.32]{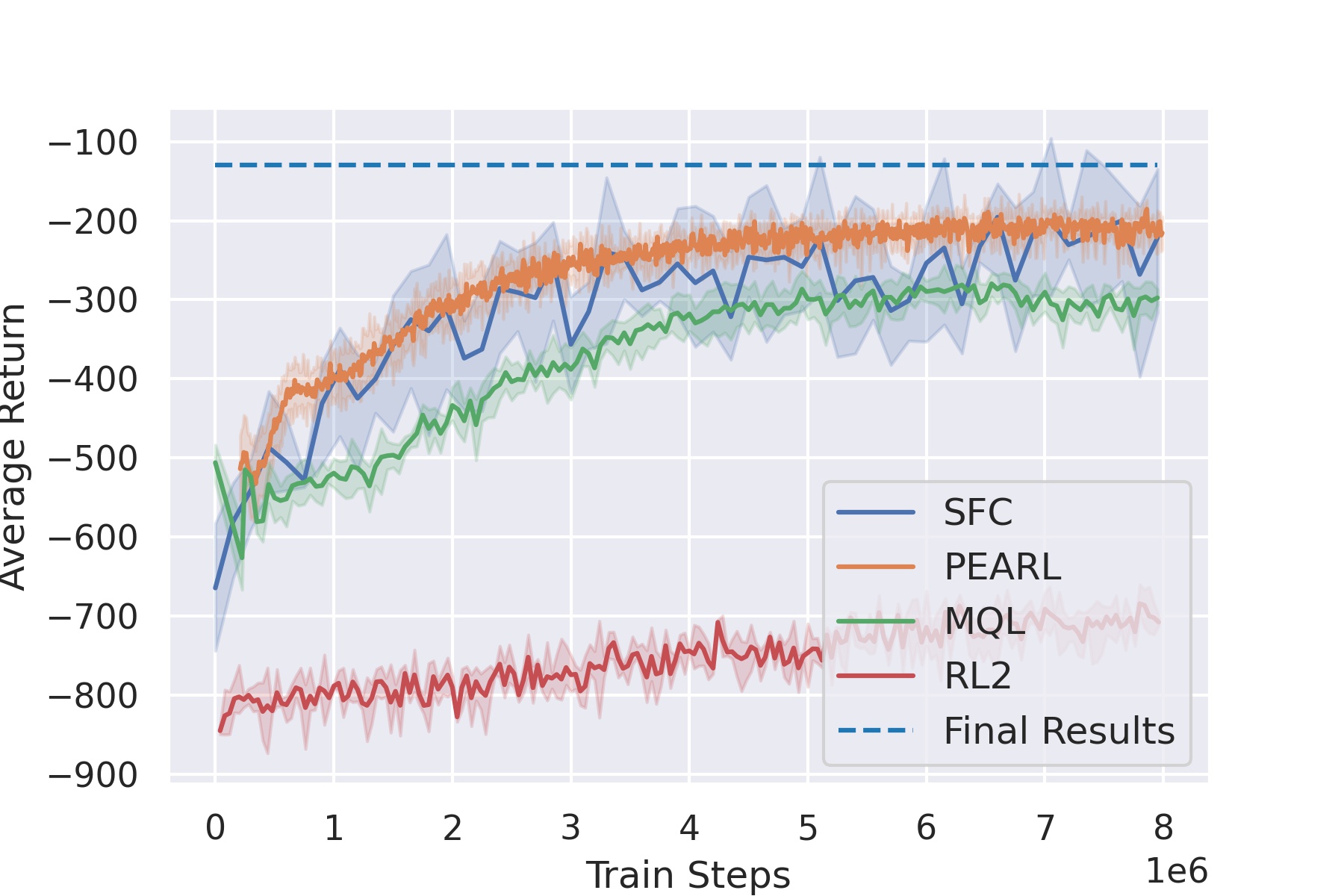}
\end{minipage}%
}%
\subfigure[HC-Vel]{
\begin{minipage}[t]{0.25\linewidth}
\centering
\includegraphics[scale=0.32]{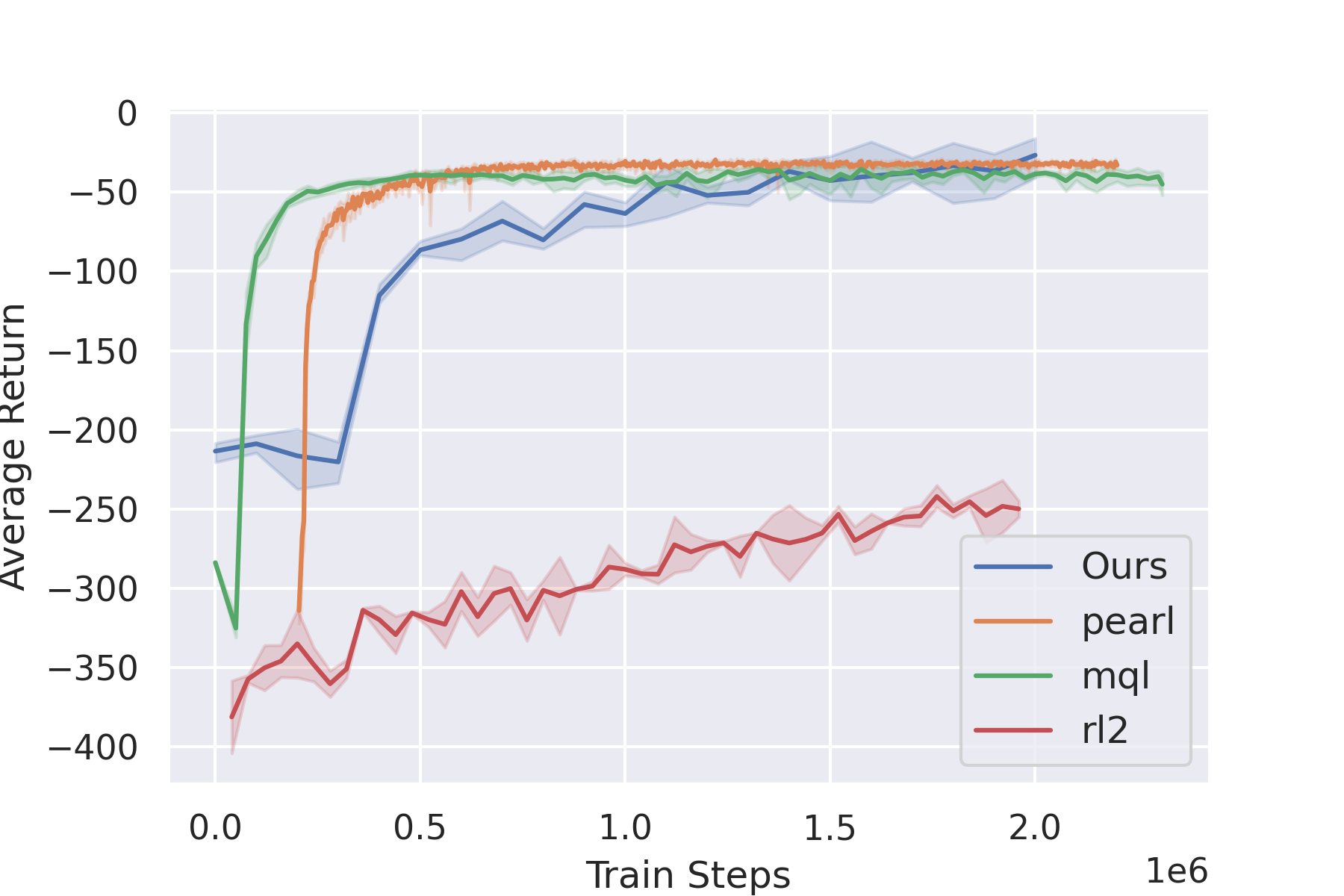}
\end{minipage}
}%
\subfigure[HC-Fwd-Back]{
\begin{minipage}[t]{0.25\linewidth}
\centering
\includegraphics[scale=0.32]{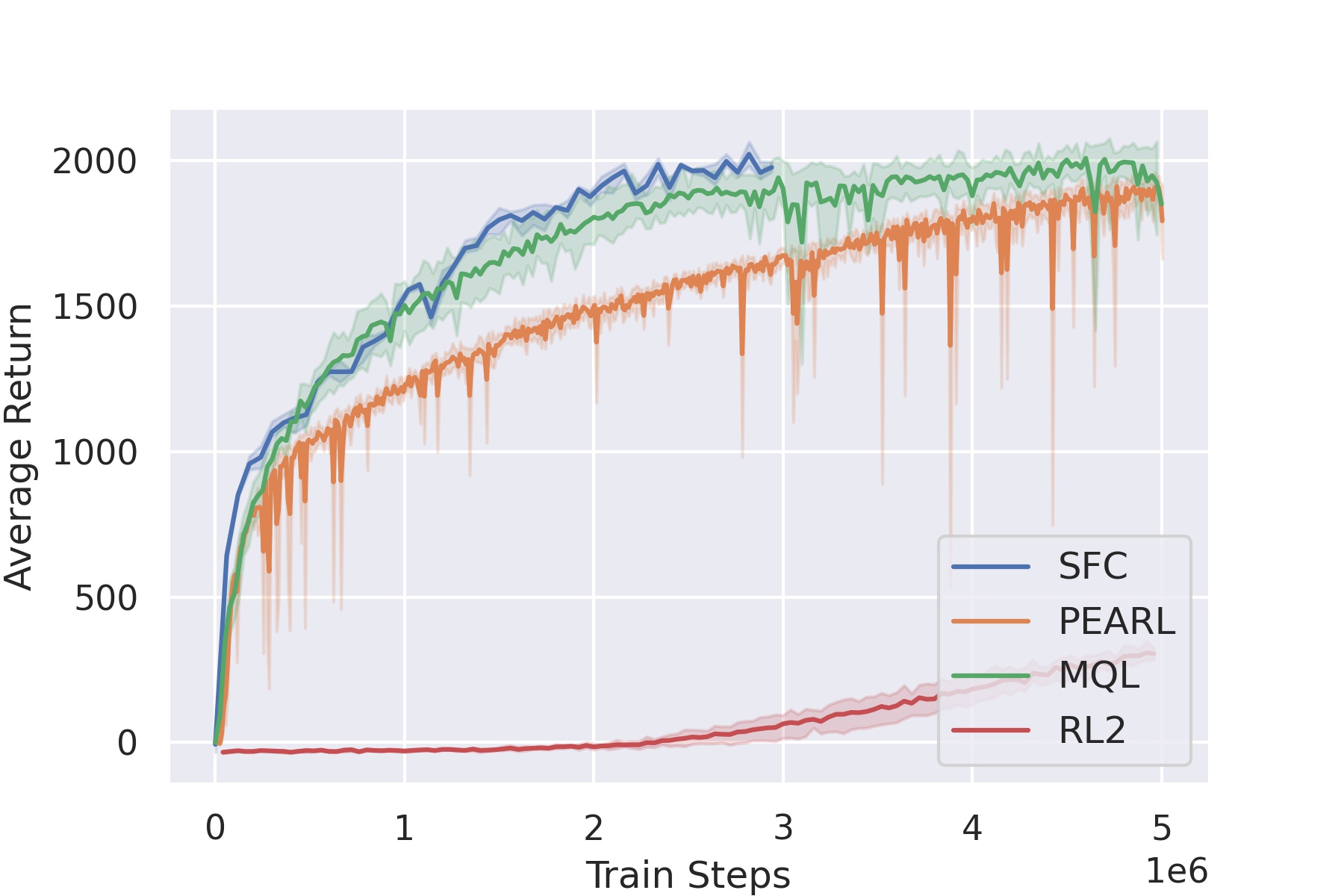}
\end{minipage}
}%
\centering
\caption{ Results of overall performance comparing our method with other meta-RL algorithms on continuous-control domains. } \label{fig:res1}
\end{figure*}

\begin{figure*}[t]
\centering
\subfigure[Ant-Fwd-Back]{
\begin{minipage}[t]{0.25\linewidth}
\centering
\includegraphics[scale=0.32]{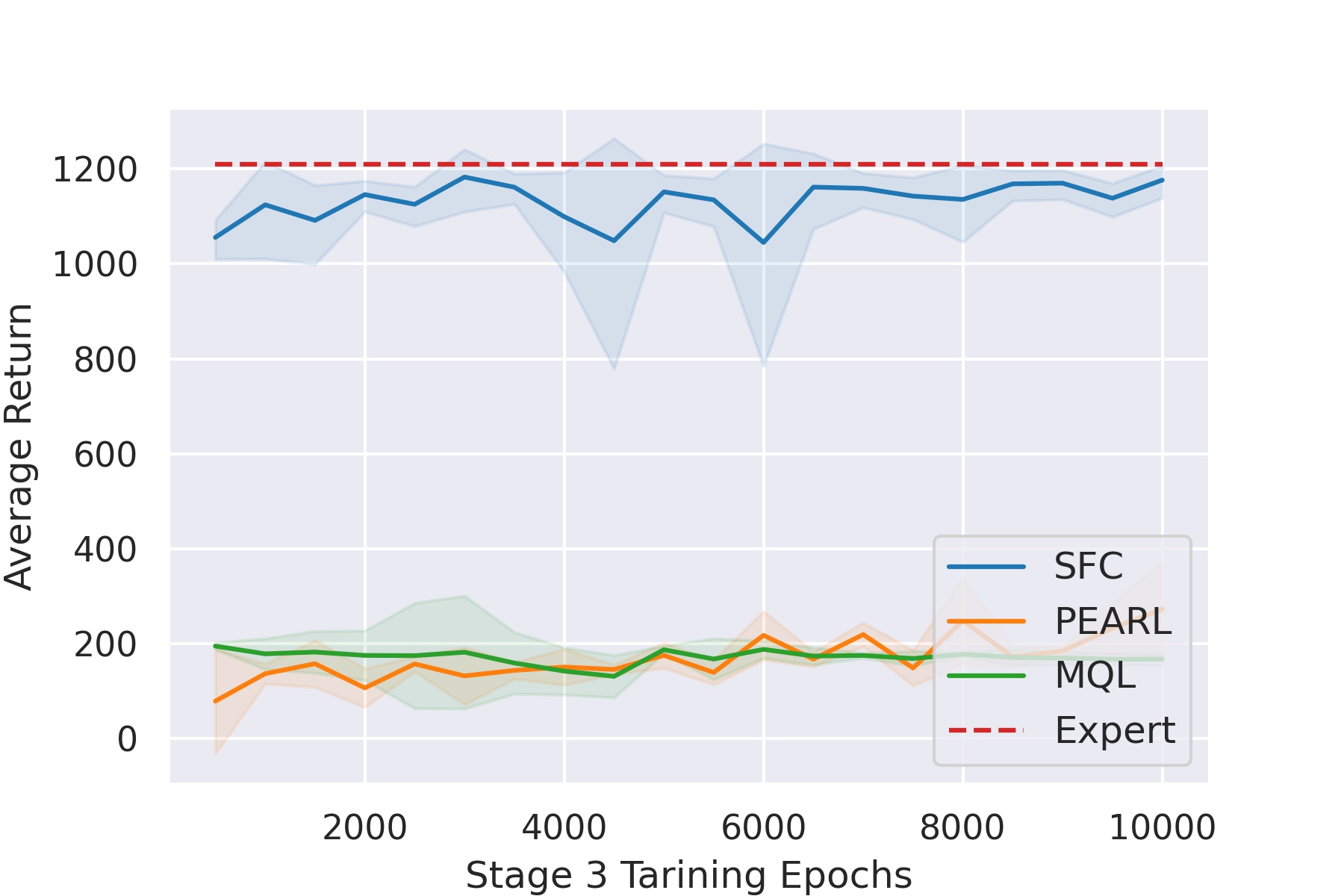}
\end{minipage}%
}%
\subfigure[Ant-Goal-2D]{
\begin{minipage}[t]{0.25\linewidth}
\centering
\includegraphics[scale=0.32]{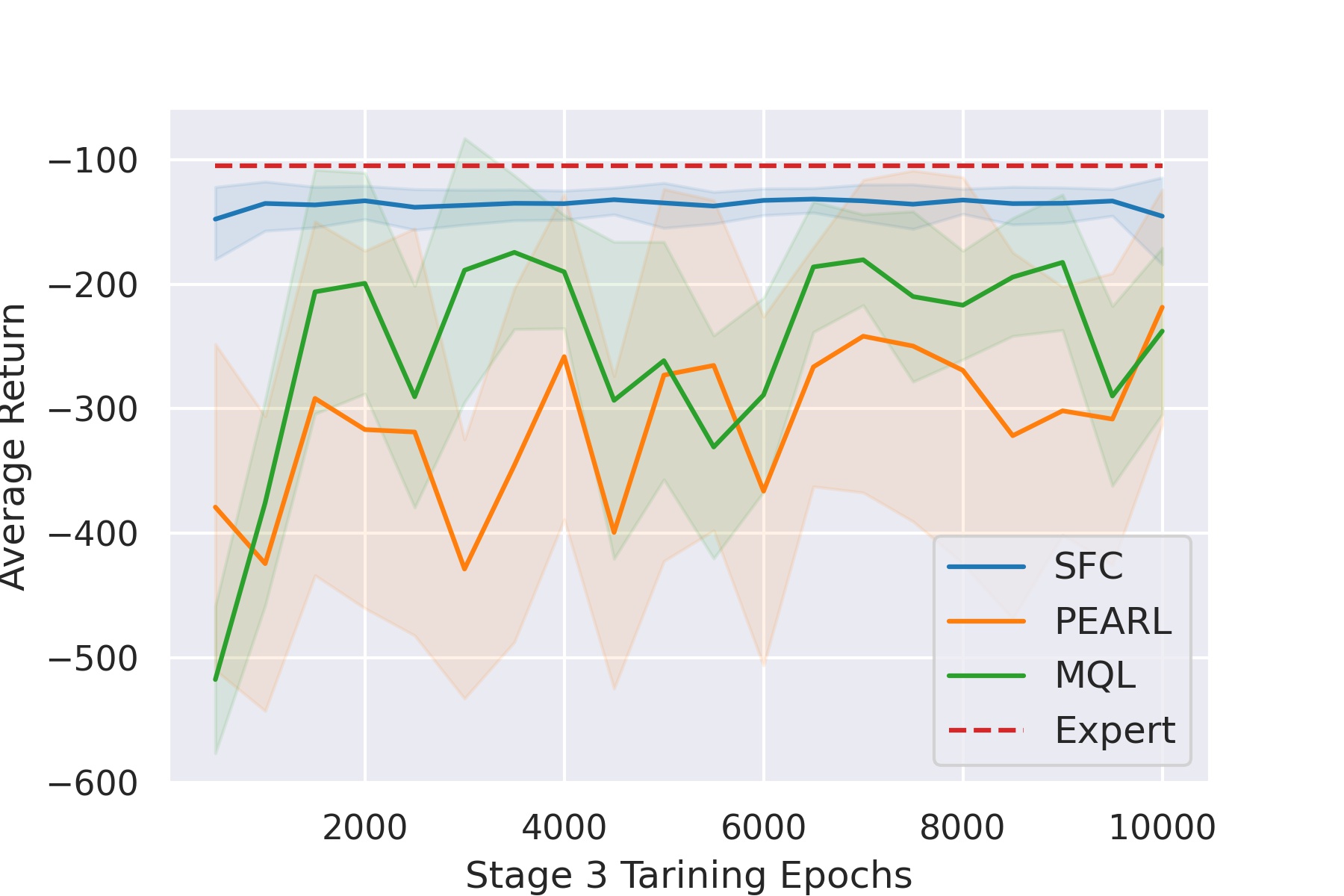}
\end{minipage}%
}%
\subfigure[HC-Vel]{
\begin{minipage}[t]{0.25\linewidth}
\centering
\includegraphics[scale=0.32]{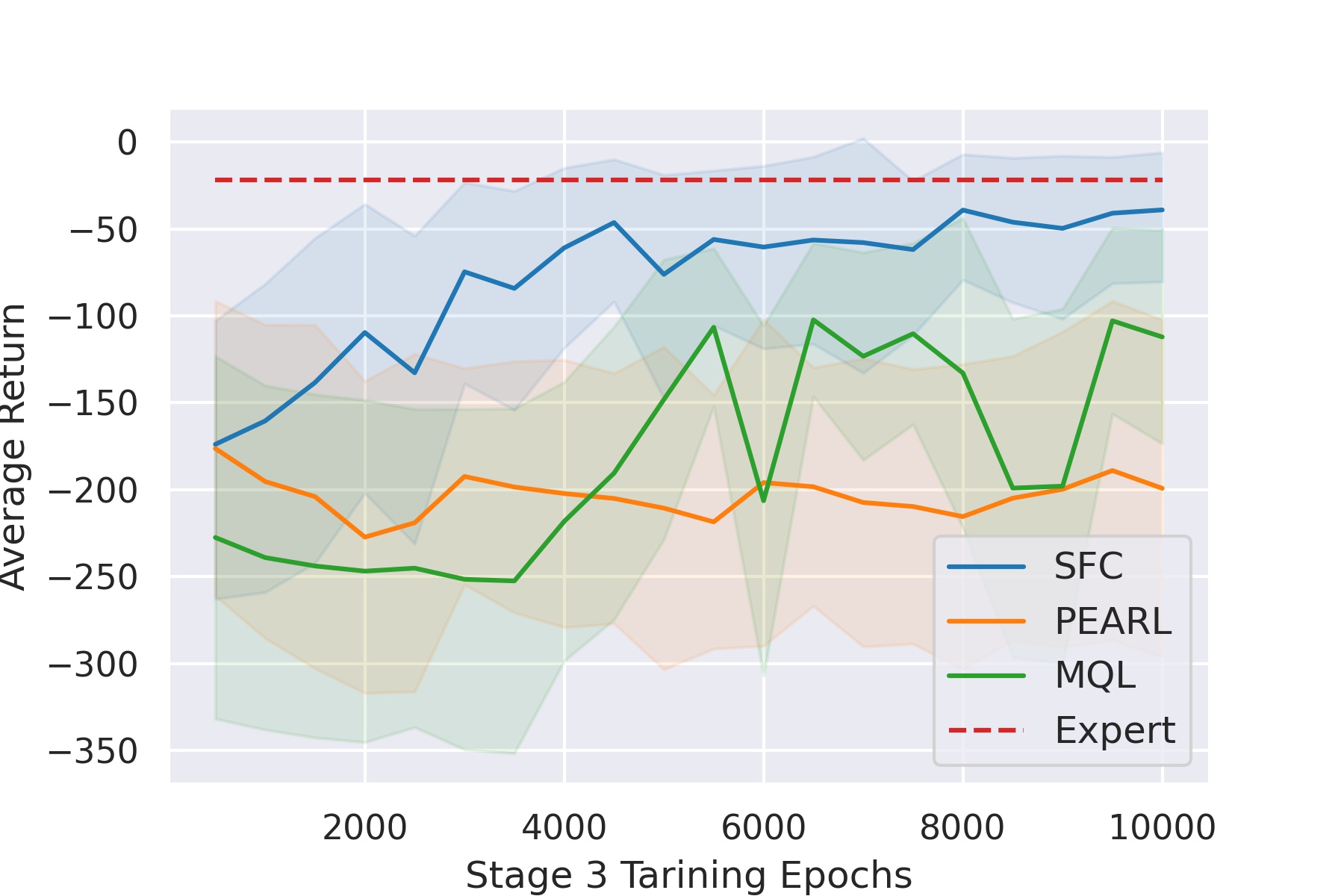}
\end{minipage}
}%
\subfigure[HC-Fwd-Back]{
\begin{minipage}[t]{0.25\linewidth}
\centering
\includegraphics[scale=0.32]{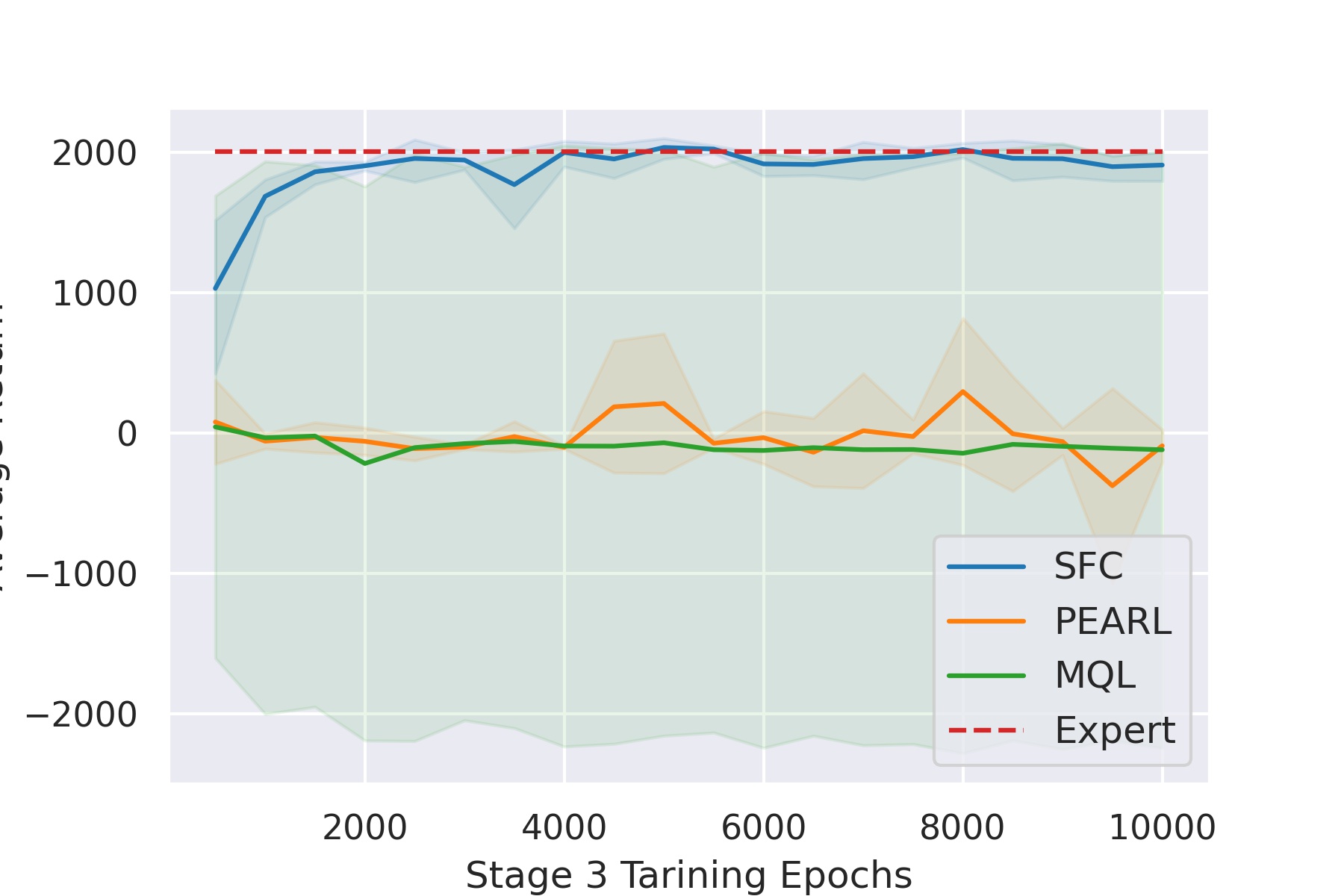}
\end{minipage}
}%
\centering
\caption{ Results of performance using different encoders in the third stage of our algorithm. } \label{fig:res2}
\end{figure*}

\begin{figure}[t]
\centering
\includegraphics[scale=0.5]{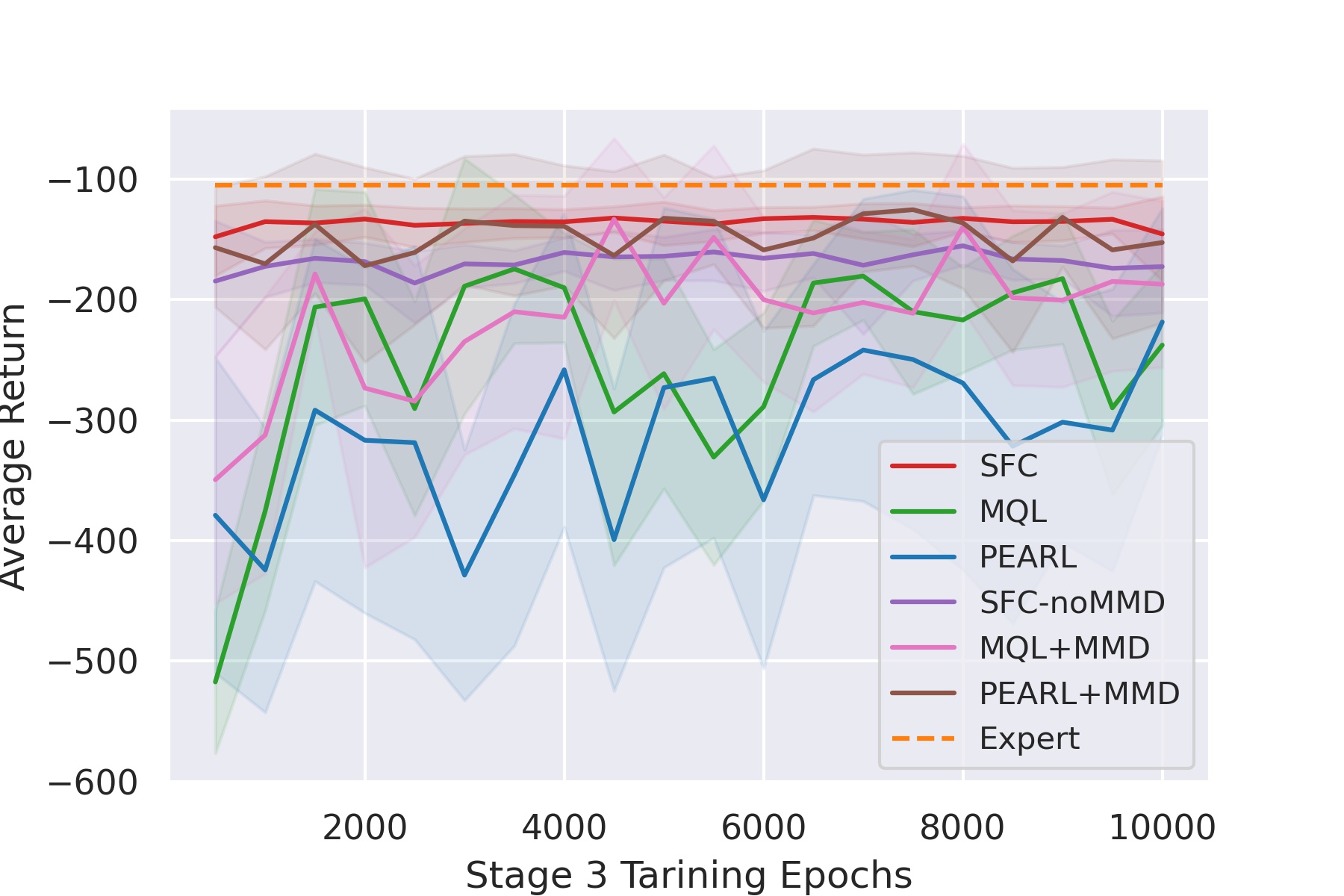}
\caption{Ablation results on the MMD loss.}
\label{fig:3}
\end{figure}

\begin{figure*}[t]
\centering

\subfigure[PEARL]{
\begin{minipage}[t]{0.3\linewidth}
\centering
\includegraphics[scale=0.4]{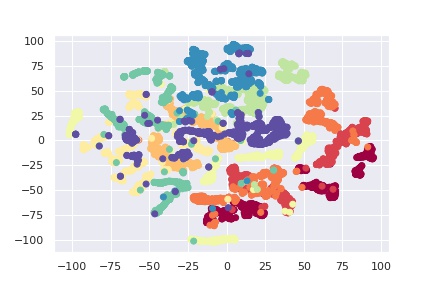}
\end{minipage}%
}%
\subfigure[SF Context]{
\begin{minipage}[t]{0.3\linewidth}
\centering
\includegraphics[scale=0.4]{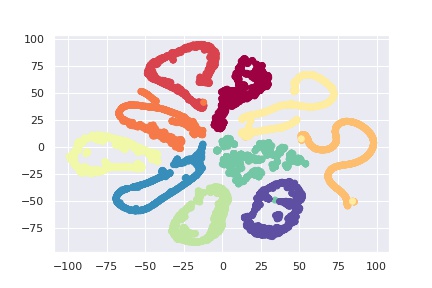}
\end{minipage}%
}%
\subfigure[SF ]{
\begin{minipage}[t]{0.3\linewidth}
\centering
\includegraphics[scale=0.4]{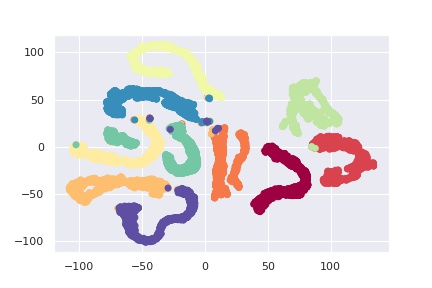}
\end{minipage}%
}%

\centering
\caption{Visualization of contexts in the embedding space (Different colors represent contexts from different tasks).} \label{fig:res4}
\end{figure*}

We conducted experiments and evaluate the performance of our method in the following aspects. Firstly, we test the overall method in several meta-RL benchmark problems and compare it with state-of-the-art approaches especially in terms of sample efficiency. Secondly, we compare the data efficiency of different encoders using a fixed dataset. Here, we want to confirm that our encoder is more efficient than other existing encoders. Finally, we visualize the results of the different context encoders to show that our method is able to extract effective and reasonable context information. This is helpful to get a better understanding of the insight of our method.

\subsection{Results on Overall Performance}


We tested our method called SFC in the four continuous control environments (i.e., Ant-Fwd-Back, Ant-Goal, HalfCheetah(HC)-Vel, and  HC-Fwd-Back), focusing on robotic locomotion using the Mujoco \cite{todorov2012mujoco} physics engine. Note that these domains are commonly used in the community to benchmark meta-RL methods \cite{finn2017model,fakoor2019meta,rakelly2019efficient}. Specifically, in Ant-Fwd-Back and HC-Fwd-Back, tasks are defined based on the target directions. Here, the agents running at maximum speed in the target direction get the maximize returns. In Ant-Goal, the agents need to reach some designated destination to complete their tasks, defined by different goal locations. In HC-Vel, tasks are defined by some constant velocities that the agents should achieve.

We compared with the three leading meta-RL algorithms:  RL$^2$ \cite{duan2016rl},  PEARL \cite{rakelly2019efficient} and MQL \cite{fakoor2019meta}. For fair comparison, we strictly followed the experimental setup of the compared algorithms as described in their papers when testing our approach for adapting to new tasks. For instance, the horizon length for all tasks is set to 200, and 3 random seeds are used to average the returns of each verification task. In order to ensure a fair comparison, we reserve the policies every fixed time step in the first stage of our algorithm and train these policies following the second and third stages to obtain a policy which is used to compared with the benchmark algorithms.

The overall results are summarized in Figure \ref{fig:res1}. As we can see from the figure, although we use a single-task RL method for training separately in different tasks, our method is still competitive to the leading meta-RL algorithms with a similar amount of environment interactions. This is mainly because the context encoder used in other meta-RL is less efficient. Therefore, they require more interactions to train the policy in order to distinguish between different tasks. In contrast, we decompose environmental dynamics and rewards for different tasks through our SF network. This information can be encoded in the policy through SF and reward weights. This makes our algorithm more effective to adapt to different tasks. In other words, the context variable proposed by our method is more efficient for reasoning new tasks and learning from the experiences.

\subsection{Results on Different Context Encoders}

Here, we tested our method using different encoders: 1) The inference network used in PEARL \cite{rakelly2019efficient}, 2) The deterministic context encoder used in MQL \cite{fakoor2019meta}. Specifically, we replaced our SF based context encoder with the encoders used in the existing meta-RL methods to train the policy in the third stage of our algorithm.

As shown in Figure \ref{fig:res2}, when only a small amount of data is offered (e.g., 10000 expert transitions in the data set of each task), the context variable in PEARL and MQL has difficulty to effectively infer the task. This still happens even for some tasks that are relatively simple. For example, in Ant-Fwd-Back and HalfCheetah-Fwd-Back, the differences between tasks are only to make the robot move in opposite directions. This is because the transitions collected by both tasks are very similar. It causes issues for other encoders to correctly infer the tasks. In Figure \ref{fig:res2}(d), although the agent using the MQL context encoder has learned how to walk, it cannot effectively distinguish the current task from others.

We also conducted ablation experiments on the MMD loss with Ant goal tasks as shown in Figure \ref{fig:3}. Specifically, we tested the following encoders for ablation: 1) The inference network used in PEARL with the MMD loss added, 2) The deterministic context encoder used in MQL with the MMD loss added, and 3) our SF context encoder without the MMD loss. As we can see from the figure, by using MMD loss to keep the context variables of different tasks away from each other, the agent can perform task inference more effectively. Additionally, we found that our algorithm still has better performance even without using the MMD loss in the third stage. This also confirms that our SF network has better performance in distinguishing different tasks by decomposing the environment dynamics and reward structure.

\subsection{Visualization of Context Variables}

Here, we collected trajectories by running the learned policies to adapt to ten randomly sampled tasks for testing. With those trajectories, we visualized the outputs of our SF network and the context encoder network, comparing with the PEARL inference networks, using t-SNE \cite{maaten2008visualizing}.

Figure \ref{fig:res4} shows the visualization results. As we can see, through the SF network, different trajectories can be distinguished from each other. This shows that the SF network can extract high-quality task-specific information from the environment. Furthermore, SFs can make the context variable more efficient combined with the MMD loss. In contrast, the results generated by the PEARL encoder are quite noisy. This also shows that our context encoder is more efficient than the one used by PEARL.

\section{Related Work}

This section briefly reviews the previous work on meta RL and successor features that are related to our method.

\subsection{Meta-RL}

Note that the main goal of meta-RL is to learn a policy that can quickly adapt to new tasks. Gradient-based meta-RL methods \cite{finn2017model,gupta2018meta,liu2019taming,stadie2018some} train a model that is expected to have a good network initialization so that different tasks can be learned quickly with policy gradient methods. Recurrent or recursive meta-RL methods \cite{duan2016rl,wang2016learning,mishra2017simple} use recurrent architectures to learn latent representations of the online experiences, and put it into policy to made themselves generalizable.

By following this track of research, context-based meta-RL methods are proposed to meta-learn from off-policy data by leveraging context. The key challenge is how to obtain an effective context variable, which motivates our work.
\citet{rakelly2019efficient} propose PEARL leverage a context inference network to adapt to a new environment with a small number of trajectories. MQL \cite{fakoor2019meta} shows that vanilla RL algorithms combined with a GRU context encoder can perform comparably to PEARL. CCM \cite{fu2020towards} introduces a method of contrast learning to make the context variable distance between different tasks farther, which improves the quality of potential context. MBML \cite{li2019multi} has improved the task inference ability by proposing a novel application of reward relabeling and triplet loss. In contrast, our method uses the successor feature framework  to directly decompose the reward using the dataset of different tasks, which is more stable and efficient as shown in our experiments.

\subsection{Successor Features}

\citet{dayan1993improving} introduced successor representations (SR) as an approach for improving generalization in temporal-difference methods. \citet{barreto2017successor} generalized it to a function approximation setting as known as successor features (SF). Due to its ability to separate environment dynamics and rewards in MDPs, SF has been widely used for better generalized navigation \cite{zhang2017deep} and control algorithms across similar environments and changing goals \cite{barreto2020fast}. As a representation defining state generalization based on the similarity of successor states, SR can be used as the count of reached states to encourage exploration \cite{machado2020count}.

Note that the SF does not only contain policy information but also environmental dynamics. Given that the reward weights only contain the reward information of the MDP, the SF and reward weights output from the SF network have more information than the input transitions. As our experimental results show, the context encoder with the SF network indeed produces more efficient context variables.

\section{Conclusion}

In this paper, we proposed a novel meta-RL algorithm. Based on the idea that the successor features framework can decouple environmental dynamics and reward structure, we put successor features and reward weights into the context encoder. By doing so, the agent can better adapt to new tasks. Our experimental results show that given a limited dataset, the performance of the encoder we designed is significantly better than the currently leading meta-RL methods. By using this efficient context encoder, we can first train for different tasks and then use a multi-task policy to efficiently integrate the policies we have learned before. Furthermore, we advance the state-of-the-art of meta RL and offer new ideas for combining different policies from previously learned tasks.


\bibliographystyle{named}
{\small \bibliography{ijcai22}}

\begin{thebibliography}{}

\bibitem[\protect\citeauthoryear{Barreto \bgroup \em et al.\egroup
  }{2017}]{barreto2017successor}
Andr{\'e} Barreto, Will Dabney, R{\'e}mi Munos, Jonathan~J Hunt, Tom Schaul,
  Hado~P van Hasselt, and David Silver.
\newblock Successor features for transfer in reinforcement learning.
\newblock In {\em Advances in neural information processing systems}, pages
  4055--4065, 2017.

\bibitem[\protect\citeauthoryear{Barreto \bgroup \em et al.\egroup
  }{2020}]{barreto2020fast}
Andr{\'e} Barreto, Shaobo Hou, Diana Borsa, David Silver, and Doina Precup.
\newblock Fast reinforcement learning with generalized policy updates.
\newblock {\em Proceedings of the National Academy of Sciences},
  117(48):30079--30087, 2020.

\bibitem[\protect\citeauthoryear{Bengio \bgroup \em et al.\egroup
  }{1992}]{bengio1992optimization}
Samy Bengio, Yoshua Bengio, Jocelyn Cloutier, and Jan Gecsei.
\newblock On the optimization of a synaptic learning rule.
\newblock In {\em Preprints Conf. Optimality in Artificial and Biological
  Neural Networks}, volume~2, 1992.

\bibitem[\protect\citeauthoryear{Cho \bgroup \em et al.\egroup
  }{2014}]{cho2014learning}
Kyunghyun Cho, Bart Van~Merri{\"e}nboer, Caglar Gulcehre, Dzmitry Bahdanau,
  Fethi Bougares, Holger Schwenk, and Yoshua Bengio.
\newblock Learning phrase representations using rnn encoder-decoder for
  statistical machine translation.
\newblock {\em arXiv preprint arXiv:1406.1078}, 2014.

\bibitem[\protect\citeauthoryear{Dayan}{1993}]{dayan1993improving}
Peter Dayan.
\newblock Improving generalization for temporal difference learning: The
  successor representation.
\newblock {\em Neural Computation}, 5(4):613--624, 1993.

\bibitem[\protect\citeauthoryear{Duan \bgroup \em et al.\egroup
  }{2016}]{duan2016rl}
Yan Duan, John Schulman, Xi~Chen, Peter~L Bartlett, Ilya Sutskever, and Pieter
  Abbeel.
\newblock Rl2: Fast reinforcement learning via slow reinforcement learning.
\newblock {\em arXiv preprint arXiv:1611.02779}, 2016.

\bibitem[\protect\citeauthoryear{Fakoor \bgroup \em et al.\egroup
  }{2019}]{fakoor2019meta}
Rasool Fakoor, Pratik Chaudhari, Stefano Soatto, and Alexander~J Smola.
\newblock Meta-q-learning.
\newblock {\em arXiv preprint arXiv:1910.00125}, 2019.

\bibitem[\protect\citeauthoryear{Finn \bgroup \em et al.\egroup
  }{2017}]{finn2017model}
Chelsea Finn, Pieter Abbeel, and Sergey Levine.
\newblock Model-agnostic meta-learning for fast adaptation of deep networks.
\newblock {\em arXiv preprint arXiv:1703.03400}, 2017.

\bibitem[\protect\citeauthoryear{Fu \bgroup \em et al.\egroup
  }{2020}]{fu2020towards}
Haotian Fu, Hongyao Tang, Jianye Hao, Chen Chen, Xidong Feng, Dong Li, and
  Wulong Liu.
\newblock Towards effective context for meta-reinforcement learning: an
  approach based on contrastive learning.
\newblock {\em arXiv preprint arXiv:2009.13891}, 2020.

\bibitem[\protect\citeauthoryear{Fujimoto \bgroup \em et al.\egroup
  }{2018}]{fujimoto2018addressing}
Scott Fujimoto, Herke Van~Hoof, and David Meger.
\newblock Addressing function approximation error in actor-critic methods.
\newblock {\em arXiv preprint arXiv:1802.09477}, 2018.

\bibitem[\protect\citeauthoryear{Gupta \bgroup \em et al.\egroup
  }{2018}]{gupta2018meta}
Abhishek Gupta, Russell Mendonca, YuXuan Liu, Pieter Abbeel, and Sergey Levine.
\newblock Meta-reinforcement learning of structured exploration strategies.
\newblock {\em arXiv preprint arXiv:1802.07245}, 2018.

\bibitem[\protect\citeauthoryear{Lee \bgroup \em et al.\egroup
  }{2020}]{lee2020context}
Kimin Lee, Younggyo Seo, Seunghyun Lee, Honglak Lee, and Jinwoo Shin.
\newblock Context-aware dynamics model for generalization in model-based
  reinforcement learning.
\newblock {\em arXiv preprint arXiv:2005.06800}, 2020.

\bibitem[\protect\citeauthoryear{Li \bgroup \em et al.\egroup
  }{2019}]{li2019multi}
Jiachen Li, Quan Vuong, Shuang Liu, Minghua Liu, Kamil Ciosek, Henrik
  Iskov~Christensen, and Hao Su.
\newblock Multi-task batch reinforcement learning with metric learning.
\newblock {\em arXiv e-prints}, pages arXiv--1909, 2019.

\bibitem[\protect\citeauthoryear{Liu \bgroup \em et al.\egroup
  }{2019}]{liu2019taming}
Hao Liu, Richard Socher, and Caiming Xiong.
\newblock Taming maml: Efficient unbiased meta-reinforcement learning.
\newblock In {\em International Conference on Machine Learning}, pages
  4061--4071, 2019.

\bibitem[\protect\citeauthoryear{Maaten and
  Hinton}{2008}]{maaten2008visualizing}
Laurens van~der Maaten and Geoffrey Hinton.
\newblock Visualizing data using t-sne.
\newblock {\em Journal of machine learning research}, 9(Nov):2579--2605, 2008.

\bibitem[\protect\citeauthoryear{Machado \bgroup \em et al.\egroup
  }{2020}]{machado2020count}
Marlos~C Machado, Marc~G Bellemare, and Michael Bowling.
\newblock Count-based exploration with the successor representation.
\newblock In {\em Proceedings of the AAAI Conference on Artificial
  Intelligence}, pages 5125--5133, 2020.

\bibitem[\protect\citeauthoryear{Mishra \bgroup \em et al.\egroup
  }{2017}]{mishra2017simple}
Nikhil Mishra, Mostafa Rohaninejad, Xi~Chen, and Pieter Abbeel.
\newblock A simple neural attentive meta-learner.
\newblock {\em arXiv preprint arXiv:1707.03141}, 2017.

\bibitem[\protect\citeauthoryear{Mnih \bgroup \em et al.\egroup
  }{2015}]{mnih2015human}
Volodymyr Mnih, Koray Kavukcuoglu, David Silver, Andrei~A Rusu, Joel Veness,
  Marc~G Bellemare, Alex Graves, Martin Riedmiller, Andreas~K Fidjeland, Georg
  Ostrovski, et~al.
\newblock Human-level control through deep reinforcement learning.
\newblock {\em nature}, 518(7540):529--533, 2015.

\bibitem[\protect\citeauthoryear{Rakelly \bgroup \em et al.\egroup
  }{2019}]{rakelly2019efficient}
Kate Rakelly, Aurick Zhou, Chelsea Finn, Sergey Levine, and Deirdre Quillen.
\newblock Efficient off-policy meta-reinforcement learning via probabilistic
  context variables.
\newblock In {\em International conference on machine learning}, pages
  5331--5340. PMLR, 2019.

\bibitem[\protect\citeauthoryear{Rothfuss \bgroup \em et al.\egroup
  }{2018}]{rothfuss2018promp}
Jonas Rothfuss, Dennis Lee, Ignasi Clavera, Tamim Asfour, and Pieter Abbeel.
\newblock Promp: Proximal meta-policy search.
\newblock {\em arXiv preprint arXiv:1810.06784}, 2018.

\bibitem[\protect\citeauthoryear{Schmidhuber}{1987}]{schmidhuber1987evolutionary}
J{\"u}rgen Schmidhuber.
\newblock {\em Evolutionary principles in self-referential learning, or on
  learning how to learn: the meta-meta-... hook}.
\newblock PhD thesis, Technische Universit{\"a}t M{\"u}nchen, 1987.

\bibitem[\protect\citeauthoryear{Silver \bgroup \em et al.\egroup
  }{2017}]{silver2017mastering}
David Silver, Julian Schrittwieser, Karen Simonyan, Ioannis Antonoglou, Aja
  Huang, Arthur Guez, Thomas Hubert, Lucas Baker, Matthew Lai, Adrian Bolton,
  et~al.
\newblock Mastering the game of go without human knowledge.
\newblock {\em nature}, 550(7676):354--359, 2017.

\bibitem[\protect\citeauthoryear{Stadie \bgroup \em et al.\egroup
  }{2018}]{stadie2018some}
Bradly~C Stadie, Ge~Yang, Rein Houthooft, Xi~Chen, Yan Duan, Yuhuai Wu, Pieter
  Abbeel, and Ilya Sutskever.
\newblock Some considerations on learning to explore via meta-reinforcement
  learning.
\newblock {\em arXiv preprint arXiv:1803.01118}, 2018.

\bibitem[\protect\citeauthoryear{Sutton and
  Barto}{2018}]{sutton2018reinforcement}
Richard~S Sutton and Andrew~G Barto.
\newblock {\em Reinforcement learning: An introduction}.
\newblock MIT press, 2018.

\bibitem[\protect\citeauthoryear{Todorov \bgroup \em et al.\egroup
  }{2012}]{todorov2012mujoco}
Emanuel Todorov, Tom Erez, and Yuval Tassa.
\newblock Mujoco: A physics engine for model-based control.
\newblock In {\em 2012 IEEE/RSJ International Conference on Intelligent Robots
  and Systems}, pages 5026--5033. IEEE, 2012.

\bibitem[\protect\citeauthoryear{Wang \bgroup \em et al.\egroup
  }{2016}]{wang2016learning}
Jane~X Wang, Zeb Kurth-Nelson, Dhruva Tirumala, Hubert Soyer, Joel~Z Leibo,
  Remi Munos, Charles Blundell, Dharshan Kumaran, and Matt Botvinick.
\newblock Learning to reinforcement learn.
\newblock {\em arXiv preprint arXiv:1611.05763}, 2016.

\bibitem[\protect\citeauthoryear{Zhang \bgroup \em et al.\egroup
  }{2017}]{zhang2017deep}
Jingwei Zhang, Jost~Tobias Springenberg, Joschka Boedecker, and Wolfram
  Burgard.
\newblock Deep reinforcement learning with successor features for navigation
  across similar environments.
\newblock In {\em 2017 IEEE/RSJ International Conference on Intelligent Robots
  and Systems (IROS)}, pages 2371--2378. IEEE, 2017.

\end{thebibliography}

\end{document}